\documentclass[11pt,a4paper]{article}
\usepackage[hyperref]{acl2020}
\usepackage{times}
\usepackage{latexsym}
\usepackage{amsmath}
\usepackage{booktabs}
\usepackage{multirow}
\usepackage{subcaption}
\usepackage{graphicx} % DO NOT CHANGE THIS
\usepackage{comment}
\usepackage{float}
\usepackage{amsfonts}

\usepackage{microtype}

\aclfinalcopy % Uncomment this line for the final submission
\def\aclpaperid{2714} %  Enter the acl Paper ID here

\title{A Study of Non-autoregressive Model for Sequence Generation}

\author{
  Yi Ren \thanks{\quad Equal contribution.}  
  \\
  Zhejiang University \\
  \texttt{rayeren@zju.edu.cn} \\\And
  Jinglin Liu \footnotemark[1] 
  \\
  Zhejiang University \\
  \texttt{jinglinliu@zju.edu.cn} \\\And
  Xu Tan  \\
  Microsoft Research Asia \\
  \texttt{xuta@microsoft.com} \\\AND
  Zhou Zhao\thanks{\quad Corresponding author} \\
  Zhejiang University \\
  \texttt{zhaozhou@zju.edu.cn} \\\And
  Sheng Zhao \\
  Microsoft STC Asia \\
  \texttt{Sheng.Zhao@microsoft.com} \\\AND
  Tie-Yan Liu \\
  Microsoft Research Asia\\
  \texttt{tyliu@microsoft.com}
}
\date{}

\begin{document}
\maketitle
\begin{abstract}

Non-autoregressive (NAR) models generate all the tokens of a sequence in parallel, resulting in faster generation speed compared to their autoregressive (AR) counterparts but at the cost of lower accuracy. Different techniques including knowledge distillation and source-target alignment have been proposed to bridge the gap between AR and NAR models in various tasks such as neural machine translation (NMT), automatic speech recognition (ASR), and text to speech (TTS). With the help of those techniques, NAR models can catch up with the accuracy of AR models in some tasks but not in some others. In this work, we conduct a study to understand the difficulty of NAR sequence generation and try to answer: (1) Why NAR models can catch up with AR models in some tasks but not all? (2) Why techniques like knowledge distillation and source-target alignment can help NAR models. Since the main difference between AR and NAR models is that NAR models do not use dependency among target tokens while AR models do, intuitively the difficulty of NAR sequence generation heavily depends on the strongness of dependency among target tokens. To quantify such dependency, we propose an analysis model called CoMMA to characterize the difficulty of different NAR sequence generation tasks. We have several interesting findings: 1) Among the NMT, ASR and TTS tasks, ASR has the most target-token dependency while TTS has the least. 2) Knowledge distillation reduces the target-token dependency in target sequence and thus improves the accuracy of NAR models. 3) Source-target alignment constraint encourages dependency of a target token on source tokens and thus eases the training of NAR models.

\end{abstract}
\section{Introduction}
Non-autoregressive (NAR) models~\citep{oord2017parallel,gu2017non,chen2019non,ren2019fastspeech}, which generate all the tokens in a target sequence in parallel and can speed up inference, are widely explored in natural language and speech processing tasks such as neural machine translation (NMT)~\citep{gu2017non,lee2018deterministic,guo2019non,wang2019non,li2019hint,guo2019fine}, automatic speech recognition (ASR)~\citep{chen2019non} and text to speech (TTS) synthesis~\citep{oord2017parallel,ren2019fastspeech}. However, NAR models usually lead to lower accuracy than their autoregressive (AR) counterparts since the inner dependencies among the target tokens are explicitly removed. 

Several techniques have been proposed to alleviate the accuracy degradation, including 1) knowledge distillation~\citep{oord2017parallel,gu2017non,guo2019non,guo2019fine,ren2019fastspeech}, 2) imposing source-target alignment constraint with fertility~\citep{gu2017non}, word mapping~\citep{guo2019non}, attention distillation~\citep{li2019hint} and duration prediction~\citep{ren2019fastspeech}. With the help of those techniques, it is observed that NAR models can match the accuracy of AR models for some tasks~\citep{ren2019fastspeech}, but the gap still exists for some other tasks~\citep{gu2017non,chen2019non}. Therefore, several questions come out naturally: (1) Why the gap still exists for some tasks? Are some tasks more difficult for NAR generation than others? (2) Why the techniques like knowledge distillation and source-target alignment can help NAR generation? 
% and 3) other methods such as iterative refinement~\citep{lee2018deterministic}, fine-tuning from an AR model~\citep{guo2019fine}, using bag-of-words loss~\citep{shao2019minimizing}. 
%\footnote{FastSpeech~\citep{ren2019fastspeech} and ParallelWaveNet~\citep{oord2017parallel} achieve nearly comparable accuracy with their AR models.}

The main difference between AR and NAR models is that NAR models do not consider the dependency among target tokens, which is also the root cause of accuracy drop of NAR models. Thus,  to better understand NAR sequence generation and answer the above questions, we need to characterize and quantify the target-token dependency, which turns out to be non-trivial since the sequences could be of different modalities (i.e., speech or text). For this purpose, we design a novel model called COnditional Masked prediction model with Mix-Attention (CoMMA), inspired by the mix-attention in ~\citet{he2018layer} and the masked language modeling in ~\citet{devlin2018bert}: in CoMMA, (1) the prediction of one target token can attend to all the source and target tokens with mix-attention, and 2) target tokens are randomly masked with varying probabilities. CoMMA can help us to measure target-token dependency using the ratio of the attention weights on target context over that on full (both source and target) context when predicting a target token: bigger ratio, larger dependency among target tokens.

We conduct a comprehensive study in this work and obtain several interesting discoveries that can answer previous questions. First, we find that the rank of the target-token dependency among the three tasks is ASR$>$NMT$>$TTS: ASR has the largest dependency while TTS has the smallest. This finding is consistent with the accuracy gap between AR and NAR models and demonstrates the difficulty of NAR generation across tasks. Second, we replace the target sequence of original training data with the sequence generated by an AR model (i.e., through knowledge distillation) and use the new data to train CoMMA; we find that the target-token dependency is reduced. Smaller target-token dependency makes NAR training easier and thus improves the accuracy. Third, source-target alignment constraint such as explicit duration prediction~\citep{ren2019fastspeech} or implicit attention distillation~\citep{li2019hint} also reduces the target-token dependency, thus helping the training of NAR models.

%as source information is important in NAR generation, a good source-target alignment is critical for the accuracy. According to our analyses, the attention weights to source tokens are diagonal and focused in both ASR and TTS while scattered in NMT, which explains that the previous works to directly use explicit duration prediction to help the NAR model in TTS, while use implicit attention distillation to help the NAR model~\citep{li2019hint} in NMT.
%If we increase the token granularity (from BPE~\citep{sennrich2015neural} to character level) in ASR and NMT tasks, the ratio decrease. 

The main contributions of this work are as follows:
\begin{itemize}
\item We design a novel model, conditional masked prediction model with mix-attention (CoMMA), to measure the token dependency for sequence generation.
\item With CoMMA, we find that: 1)  Among the three tasks, ASR is the most difficult and TTS is the least for NAR generation; 2) both knowledge distillation and imposing source-target alignment constraint reduce the target-token dependency, and thus reduce the difficulty of training NAR models.
\end{itemize}

\section{CoMMA}
In this section, we analyze the token dependency in the target sequence with a novel conditional masked prediction model with mix-attention (CoMMA). We first introduce the design and structure of CoMMA, and then describe how to measure the target token dependency based on CoMMA.

%the attention density ratio which is used to characterize how much attention weights are focusing on the target tokens when predicting a target token.

\subsection{The Design of CoMMA}

\begin{figure*}[!thb]
	\centering
	\begin{subfigure}[h]{0.49\textwidth}
	\centering
	\includegraphics[width=\textwidth,trim={0cm 0.0cm 0cm 0cm}, clip=true]{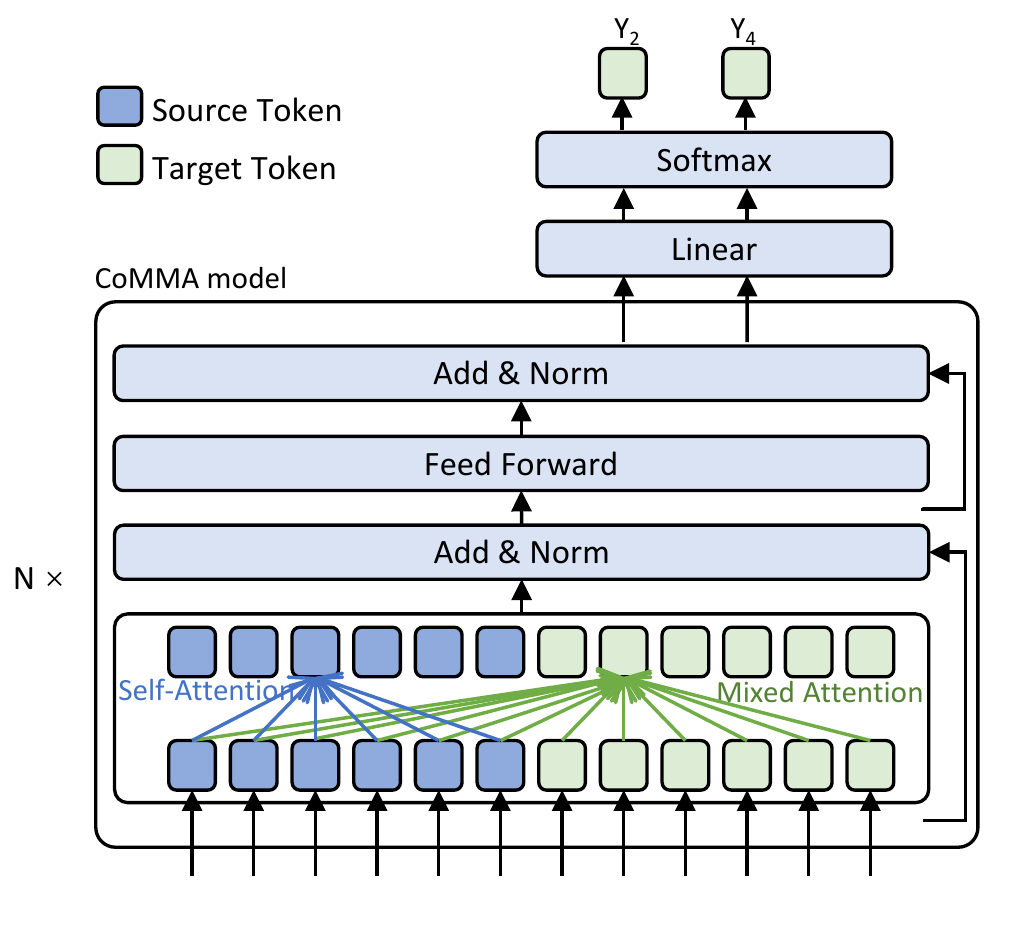}
	\caption{The main structure of CoMMA.}
	\label{simul_archi_model}
	\end{subfigure}
	\begin{subfigure}[h]{0.50\textwidth}
	\centering
	\vspace{0.1cm}
	\includegraphics[width=\textwidth,trim={0cm 0.0cm 0cm 0cm}, clip=true]{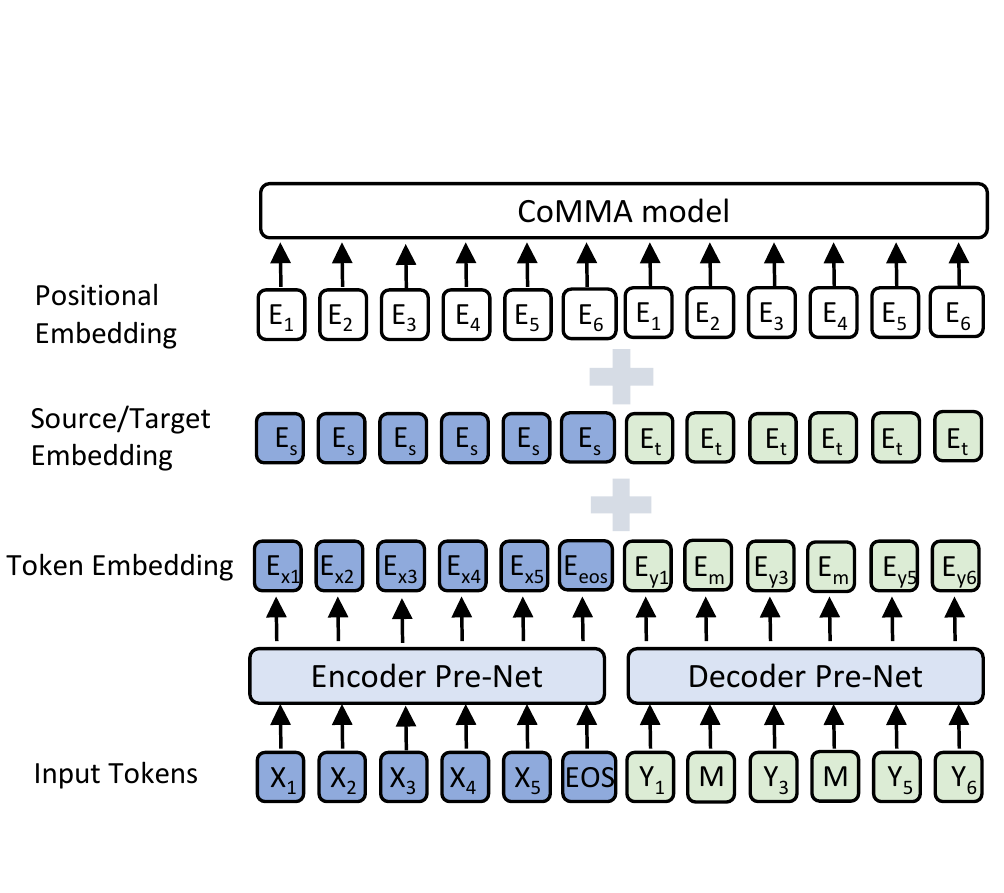}
% 	\vspace{0.1cm}
	\caption{The input module of CoMMA.}
	\label{arch_pipeline}
	\end{subfigure}
	\hspace{0.3cm}
	\caption{The architecture of conditional masked prediction model with mix-attention (CoMMA).}
	\label{fig_comma}
\end{figure*}
% \begin{figure*}[!thb]
% 	\centering
% 	\includegraphics[width=0.8\textwidth,trim={0cm 0.0cm 0cm 0cm}, clip=true]{fig/ana_model.pdf}
% 	\caption{The architecture of conditional masked prediction model with mix-attention.}
% 	\label{fig_comma}
% \end{figure*}

It is non-trivial to directly measure and compare the target token dependency in different modalities (i.e., speech or text) and different conditional source modalities (i.e., speech or text). Therefore, we have several considerations in the design of CoMMA: 1) We use masked language modeling in BERT~\citep{devlin2018bert} with source condition to train CoMMA, which can help measure the dependency on target context when predicting the current masked token. 2) In order to ensure the dependency on source and target tokens can be comparable, we use mix-attention~\citep{he2018layer} to calculate the attention weights on both source and target tokens in a single softmax function.

The model architecture of CoMMA is shown in Figure~\ref{fig_comma}. Specifically, CoMMA differs from standard Transformer~\citep{vaswani2017attention} as follows: 1) Some tokens are randomly replaced by a special mask token $\langle M \rangle$ with probability $p$, and the model is trained to predict original unmasked tokens. 2) We employ mix-attention mechanism~\citep{he2018layer} where layer $i$ in the decoder can attend to itself and the layer $i$ in the encoder at the same time and compute the attention weights in a single softmax function. We share the parameters of attention and feed-forward layer between the encoder and decoder. 3) Following~\citet{he2018layer}, we add source/target embedding to tell the model whether a token is from the source or target sequence, and also add position embedding with the positions of source and target tokens both starting from zero. 4) The encoder and decoder pre-net~\citep{shen2018natural} vary in different tasks: For TTS, encoder pre-net consists of only embedding lookup table, and decoder pre-net consists of 2-layer dense network with ReLU activation. For ASR, encoder pre-net consists of 3-layer 2D convolutional network, and decoder pre-net consists of only embedding lookup table. For NMT, both encoder and decoder pre-net consist of only embedding lookup table. 

CoMMA is designed to measure the target token dependency in a variety of sequence generations, including AR (unidirectional) generation, NAR generation, bidirectional generation or even identity copy. To this end, we vary the mask probability $p$ (the ratio of the masked tokens in the whole target tokens\footnote{Considering the continuity of the mel-spectrogram frames in speech sequence, we mask the frames by chunk, each chunk with frame size 10.}) in a uniform distribution $p\sim U(0.0, 1.0)$ when training CoMMA. In this way, $p=1$ covers NAR generation, $p=0$ covers identity copy, and in some cases, $p$ can also cover AR generation. 

%For each iteration, we concatenate the masked target sequence with the source sequence, and feed them into CoMMA to predict the original tokens in masked positions with cross-entropy loss.

\subsection{How to Measure Target Token Dependency based on CoMMA}
\label{sec_ratio}
To measure the target token dependency, we define a metric called attention density ratio $R$, which represents the ratio of the attention density (the normalized attention weights) on target context in mix-attention when predicting the target token with a well-trained CoMMA. We describe the calculation of $R$ in the following steps.

First, we define the attention density ratio $\alpha$ for a single target token $i$ as 
\begin{equation}
\small
 \alpha_{i} =  \frac{\frac{1}{N} \sum_{j=1}^{N}A_{i,j}}{ \frac{1}{N} \sum_{j=1}^{N}A_{i,j} + \frac{1}{M} \sum_{j=N+1}^{N+M}A_{i,j} },
\end{equation}
where $A_{i,j}$ denotes the attention weights from token $i$ to token $j$ in mix-attention, and $i \in [1, N]$ represents the target token while $j \in [N+1, N+M]$ represents the source token, $M$ and $N$ is the length of source and target sequence respectively, $\sum_{j=1}^{N+M}A_{i,j} = 1$. $\alpha_i$ represents the ratio of attention density on target context when predicting target token $i$.

Second, we average the attention density ratio $\alpha_i$ over all the predicted tokens (with masked probability $p$) in a sentence and get 
\begin{equation}
\small
\frac{1}{|\mathcal{M}^p|}\sum_{i \in \mathcal{M}^p}\alpha_{i},
\end{equation}
where $\mathcal{M}^p$ represents the set of masked target tokens under mask probability $p$ and $|\mathcal{M}^p|$ denotes the number of tokens in the set.

Third, for a given $p$, we calculate $R(p)$ over all test data and average them to get the final attention density ratio 
\begin{equation}
\small
R(p) = \text{Avg}(\frac{1}{|\mathcal{M}^p|}\sum_{i \in \mathcal{M}^p}\alpha_{i}).
\end{equation}
We vary $p$ and calculate $R(p)$ to measure the density ratio under different conditions, where a small $p$ represents more target context that can be leveraged and a large $p$ represents less context. In the extreme cases, $p=1$ represent NAR generation while $p=0$ represents to learn identity copy. 

\begin{table*}[!h]
\small
	\centering
	\begin{tabular}{ l | c | c | c}
		\toprule
		Task & NMT & ASR & TTS \\
		\midrule
		AR & Transformer~\citep{vaswani2017attention} &  Transformer ASR~\citep{karita2019comparative} & Transformer TTS~\citep{li2019neural} \\
		NAR & NAT~\citep{gu2017non} w/ AC & NAR-ASR~\citep{chen2019non} w/ AC & FastSpeech~\citep{ren2019fastspeech} \\
		\bottomrule
	\end{tabular}
	\vspace{0.3cm}
	\caption{The AR and NAR model we consider in each task. ``AC" means attention constraint we mentioned in Section \ref{sec:study_ac}. }
	\label{ar_nar_model}
	\vspace{-0.2cm}
\end{table*}

Given the proposed attention density ratio $R(p)$ based on CoMMA, we can measure the target token dependency of the NAR model in different tasks, which can help understand a series of important research questions, as we introduce in the following three sections.

\section{Study on the Difficulty of NAR Generation}

In this section, we aim to find out why the gap still exists for ASR and NMT tasks, while in TTS, NAR can catch up with the accuracy of AR model. We also analyze the causes of different difficulties for different tasks. We start from evaluating the accuracy gap between AR and NAR models for NMT, ASR and TTS, and then measure the token dependency based on our proposed CoMMA. 

\subsection{The Accuracy Gap}
We first train the AR and NAR models in each task and check the accuracy gap between AR and NAR models to measure the difficulty of NAR generation in each task. 

\paragraph{Configuration of AR and NAR Model}
The AR and NAR models we considered are shown in Table~\ref{ar_nar_model}, where we use Transformer as the AR models while the representative NAR models in each task. For a fair comparison, we make some modifications on the NAR models: 1) For ASR, we train a Transformer ASR first as teacher model and then constrain the attention distributions of NAR-ASR with the alignments converted by teacher attention weights, which will be introduced and discussed in Section \ref{sec:study_ac}. 2) For NMT, we constrain the KL-divergence of the encoder-to-decoder attention distributions between the AR and NAR models following \citet{li2019hint}. We also list the hyperparameters of AR and NAR models for each task in the supplementary material (Section 1).

% In order to purely measure the difficulty of each task, we make some modifications on the NAR models: 1) For NAT~\citep{gu2017non}, we remove knowledge distillation and fertility prediction, and expand the source tokens to match the length of target tokens and take them as decoder input. We call this modified model as ``NAR-NMT". 2) For FastSpeech~\citep{ren2019fastspeech}, we remove the knowledge distillation, and replace the length regulator with decoder-to-encoder attention and replace the original decoder input with positional encoding. We call this model ``NAR-TTS".  We also list the hyperparameters and training details of the AR and NAR models for each task in the supplementary material (Section 1).

\paragraph{Datasets and Evaluations for NMT, ASR and TTS } 
\label{gap_exp_settings}
We conduct experiments on IWSLT 2014 German-English (De-En) translation dataset\footnote{https://wit3.fbk.eu/mt.php?release=2014-01} for NMT, LibriTTS dataset~\citep{zen2019libritts} for ASR and LJSpeech dataset~\citep{itolj} for TTS. For speech data, we transform the raw audio into mel-spectrograms following~\citet{shen2018natural} with 50 ms frame size and 12.5 ms hop size. For text data, we tokenize sentences with moses tokenizer\footnote{https://github.com/moses-smt/mosesdecoder/blob/mast er/scripts/tokenizer/tokenizer.perl} and then segment into subword symbols using Byte Pair Encoding (BPE)~\cite{sennrich2015neural} for subword-level analysis, and convert the text sequence into phoneme sequence with grapheme-to-phoneme conversion~\citep{sun2019token} for phoneme-level analysis. We use BPE for NMT and ASR, while phoneme for TTS by default unless otherwise stated. We train all models on 2 NVIDIA 2080Ti GPUs using Adam optimizer with $\beta_{1}= 0.9$, $\beta_{2} = 0.98$, $\varepsilon = 10^{-9}$ and following the same learning rate schedule in \citep{vaswani2017attention}. 
 
For ASR, we evaluate word error rate (WER) on test-clean set in LibriTTS dataset. For NMT, we evaluate the BLEU score on IWSLT 2014 De-En test set. For TTS, we randomly split the LJSpeech dataset into 3 sets: 12500 samples for training, 300 samples for validation and 300 samples for testing, and then evaluate the mean opinion score (MOS) on the test set to measure the audio quality. The output mel-spectrograms of TTS model are transformed into audio samples using the pretrained WaveGlow~\citep{prenger2019waveglow}. Each audio is listened by at least 20 testers, who are all native English speakers.

\begin{table}[!h]
\small
	\centering
	\begin{tabular}{ l | c | c }
		\toprule
		Task & Model & Accuracy \\
		\midrule
		\multirow{2}{*}{NMT (BLEU/WER) } & \textit{Transformer} & 33.90/47.18 \\
		             & \textit{NAT} & 27.12/54.90   \\
		\midrule 
		\multirow{2}{*}{ASR (BLEU/WER)}  & \textit{Transformer ASR} & 66.60/20.10 \\
         & \textit{NAR-ASR} & 39.23/36.20   \\
         \midrule 
		\multirow{2}{*}{TTS (MOS)}  & \textit{Transformer TTS} & 3.82 $\pm$ 0.08 \\
		& \textit{FastSpeech} & 3.79 $\pm$ 0.12 \\
		\bottomrule
	\end{tabular}
	\caption{The accuracy gap between NAR and AR models.}
	\label{tab_gap_results}
\end{table}

\paragraph{Results of Accuracy Gap}
The accuracies of the AR and NAR models in each task are shown in Table~\ref{tab_gap_results}.
It can be seen that NAR model can match the accuracy of AR model gap in TTS, while the gap still exists in ASR and NMT. We calculate both the WER and BLEU metrics in ASR and NMT for better comparison. It can be seen that ASR has a larger gap than NMT. Larger accuracy gap may indicate more difficult for NAR generation in this task. Next, we try to understand what factors influence difficulties among different tasks.

\subsection{The Token Dependency}
In the last subsection, we analyze the difficulty of NAR models from the perspective of the accuracy gap. In this subsection, we try to find evidence from the target token dependency, which is supposed to be consistent with the accuracy gap to measure the task difficulty. 
\paragraph{Configuration of CoMMA}
We train CoMMA with the same configuration on NMT, ASR and TTS: the hidden size and the feed-forward hidden size and the number of layers are set to 512, 1024 and 6 respectively. We list other hyperparameters of CoMMA in the supplementary material (Section 2). We also use the same datasets for each task as described in Section~\ref{gap_exp_settings} to train CoMMA.

\begin{figure}[h]
	\centering
	\includegraphics[width=0.48\textwidth]{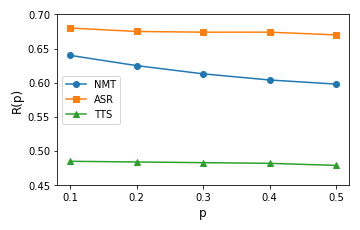}
	\caption{Attention density ratio $R(p)$ under different $p$ in different tasks for performance gap analysis.}
	\label{Adp_nokd}
\end{figure}

\paragraph{Results of Token Dependency}
We use the attention density ratio calculated from CoMMA (as described in Section~\ref{sec_ratio}) to measure the target token dependency and show the results in Figure~\ref{Adp_nokd}. It can be seen that the rank of attention density ratio $R(p)$ is ASR$>$NMT$>$TTS for all $p$. Considering that $R(p)$ measures how much context information from target side is needed to generate a target token, we can see that ASR has more dependency on the target context and less on the source context, while TTS is the opposite, which is consistent with the accuracy gap between AR and NAR models as we described in Section~\ref{gap_exp_settings}. 

As we vary $p$ from 0.1 to 0.5, $R(p)$ decreases for all tasks since more tokens in the target side are masked. We also find that $R(p)$ in NMT decreases quicker than the other two tasks, which indicates that NMT is good at learning from source context when less context information can be leveraged from the target side while $R(p)$ in ASR decreases little. This can also explain why NAR in NMT achieves less gap than ASR. 

% \begin{table}[h]
% 	\centering
% 	\begin{tabular}{ l | c | c | c }
% 	\toprule
% 	Tasks   &  $p$=0.1   & $p$=0.3    & $p$=0.5 \\
% 		\midrule
% 		\textit{NMT}       & 0.640  & 0.613 & 0.598 \\
% 		\textit{ASR}       & 0.680  & 0.674 & 0.670 \\
% 		\textit{TTS}   & 0.485  & 0.483 & 0.479 \\
% 		\bottomrule
% 	\end{tabular}
% 	\caption{Attention density ratio $R(p)$ under different $p$ in different tasks for performance gap analysis.}
% 	\label{tab_comma_gap_results}
% \end{table}

\section{Study on Knowledge Distillation}

In the current and next sections, we investigate why some techniques can help NAR generation from the aspect of target token dependency. We only analyze knowledge distillation and attention alignment techniques which are widely used in NAR, but we believe our analysis method can be applied to other NAR techniques, such as iterative refinement~\citep{lee2018deterministic}, fine-tuning from an AR model~\citep{guo2019fine} and so on.   

Most existing NAR models~\citep{oord2017parallel,gu2017non,wang2019non,guo2019non,guo2019fine,ren2019fastspeech} rely on the technique of knowledge distillation, which generates the new target sequence given original source sequence from a pre-trained AR model and trains the NAR model for better accuracy. In this section, we first conduct experiments to verify the accuracy improvements of knowledge distillation. Next, based on our proposed CoMMA, we analyze why knowledge distillation could help NAR models.

\subsection{The Effectiveness of Knowledge Distillation}

\paragraph{Knowledge Distillation for NAR Models} 
Given a well-trained AR model $\theta_T$ and source sequence $x \in \mathcal{X}$ from the original training data, a new target sequence can be generated through
\begin{equation}
 y' \sim P(y|x;\theta_T).
\label{kd_generate}
\end{equation}
We can use beam search for NMT and ASR and greedy search for TTS to generate $y'$. Given the set of generated sequence pairs $(\mathcal{X}, \mathcal{Y}')$, we train the NAR models with negative log-likelihood loss
\begin{equation}
\mathcal{L} ((\mathcal{X}, \mathcal{Y}'); \theta) = -\sum_{(x, y') \in (\mathcal{X}, \mathcal{Y}')} \log P(y'|x;\theta), 
\label{kd_train}
\end{equation}
where $\theta$ is the parameters set of the NAR model.

\begin{table}[!h]
\small
	\centering
	\begin{tabular}{ l | c | c }
		\toprule
		Task & Model & Accuracy \\
		\midrule
		\multirow{3}{*}{NMT (BLEU) } & \textit{Transformer} & 33.90 \\
		\cmidrule{2-3}
		             & \textit{NAT} & 27.12   \\
		             & \textit{NAT w/o KD} & 21.79  \\
         \midrule 
		\multirow{3}{*}{TTS (MOS)}  & \textit{Transformer TTS} & 3.82 $\pm$ 0.08 \\
		\cmidrule{2-3}
		            & \textit{FastSpeech} & 3.79 $\pm$ 0.12 \\
		            & \textit{FastSpeech w/o KD} & 3.58 $\pm$ 0.13 \\
		\bottomrule
	\end{tabular}
	\caption{The comparison between NAR models with and without knowledge distillation.}
	\label{tab_kd_results}
\end{table}

\paragraph{Experimental Results}
We only conducted knowledge distillation on NMT and TTS since there is no previous works on ASR yet. We train the NAR models in NMT and TTS with raw target token sequence instead of teacher outputs and compare the results with that in Table \ref{tab_gap_results}. The accuracy improvements of knowledge distillation are shown in Table~\ref{tab_kd_results}. It can be seen that knowledge distillation can boost the accuracy of NAR in NMT and TTS, which is consistent with the previous works.

% We choose NAR-NMT and NAR-TTS as the NAR models and use same configuration and datasets as in Section \ref{gap_exp_settings}. We first train an AR model as the teacher, generate the target sequence given source sentence in the original training set according to Equation~\ref{kd_generate}, and then train the NAR student model according to Equation~\ref{kd_train}. 

\subsection{Why Knowledge Distillation Works}
Recently, \citet{zhou2019understanding} find that knowledge distillation can reduce the complexity of data sets and help NAT to better model the variations in the output data.  However, this explanation is reasonable on its own, but mainly from the perspective of data level and is not easy to understand. In this subsection, we analyze knowledge distillation from a more understandable and intuitive perspective, by observing the change of the token dependency based on our proposed CoMMA.

\begin{figure}[h]
	\centering
	\includegraphics[width=0.48\textwidth]{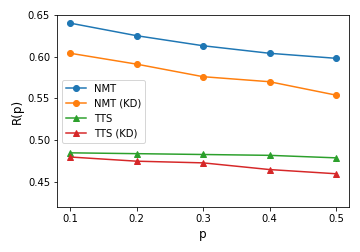}
	\caption{Attention density ratio $R(p)$ for NMT and TTS tasks under different $p$ with and without knowledge distillation, where ``KD" means knowledge distillation.}
	\label{Adp_kd}
	\vspace{-0.2cm}
\end{figure}

We measure the target token dependency by training CoMMA with the original training data and new data generated through knowledge distillation, respectively. The results are shown in Figure~\ref{Adp_kd}. It can be seen that knowledge distillation can decrease the attention density ratio $R(p)$ on both tasks, indicating that knowledge distillation can reduce the dependency on the target-side context when predicting a target token, which can be helpful for NAT model training.
% Table \ref{tab_adp_kd_results}
% \begin{table}[h]
% 	\centering
% 	\begin{tabular}{ l | c | c | c  }
% 		\toprule
% 		Tasks          &  p=0.1  & p=0.3    & p=0.5 \\
% 		\midrule
% 		\textit{NMT (BPE)}       & 0.640 & 0.613 & 0.598 \\
% 		\textit{NMT (BPE+KD)}    & 0.604 & 0.576 & 0.554 \\
% 		\midrule     
% 		\textit{TTS (phoneme)}   & 0.485 & 0.483 & 0.479 \\
% 		\textit{TTS (phoneme+KD)}& 0.480 & 0.473 & 0.460 \\
% 		\bottomrule
% 	\end{tabular}
% 	\caption{Attention density ratio $R$ for different tasks on different $p$}
% 	\label{tab_adp_kd_results}
% \end{table}

\section{Study on Alignment Constraint}
\label{sec:study_ac}

Without the help of target context, NAR models usually suffer from ambiguous attention to the source context, which affects the accuracy. Recently, many works have proposed a variety of approaches to help with the source-target alignment of NAR models, which can improve the estimation of the soft alignment in attention mechanism model. For example, \citet{li2019hint} constrain the KL-divergence of the encoder-to-decoder attention distributions between the AR and NAR models. \citet{gu2017non} predict the fertility of the source tokens to approximate the alignments between target sequence and source sequence. \citet{guo2019non} convert the source token to target token with phrase table or embedding mapping for alignments. \citet{ren2019fastspeech} predict the duration (the number of mel-spectrograms) of each phoneme.

% which can be classified into two categories: 1) Implicit alignment constraint, which adds constraint on the attention of NAR model in training stage. For example,  2) Explicit alignment constraint, which calculates the number of target tokens that corresponds to each source token and then expand the source sequence to match the length of target sequence. For example, \citet{gu2017non} predict the fertility of the source tokens, and \citet{ren2019fastspeech} predict the duration (the number of mel-spectrograms) of each phoneme. In both the implicit and explicit methods, an AR model is usually leveraged to help on the alignment constraint, e.g., \citet{li2019hint} leverage an AR model for attention distillation while \citet{ren2019fastspeech} leverage an AR model to extract the duration information to train the duration predictor.

In this section, we first study the effectiveness of alignment constraint for NAR models, and then analyze why alignment constraint can help the NAR models by observing the changes of token dependency based on our proposed CoMMA.

\subsection{The Effectiveness of Alignment Constraint}
\paragraph{Alignment Constraint for NAR Models}

We choose the attention constraint mechanism which is commonly used based on previous works for each task. 

For NMT, we follow ~\citet{li2019hint} to minimize the KL-divergence between the attention distributions of AR and NAR model as follow: 
\begin{equation}
    \mathcal{L}_{ac} = \frac{1}{N}\sum_{i=1}^{N}{D_{KL}(A'_{i}||A_{i})},
    \label{eqn_iac}
\end{equation}
where $A'_{i}$ and $A_{i}$ denote the source-target attention weights from the AR teacher model and NAR student model respectively. $A', A \subset \mathbb{R}^{N \times M} $ where $N$ and $M$ are the number of tokens in the target and source sequence. 

For TTS, we follow \citet{ren2019fastspeech} to extract the encoder-to-decoder attention alignments from the well-trained AR teacher model and convert them to phoneme duration sequence, and then train the duration predictor to expand the hidden of the source sequence to match the length of target sequence.

For ASR, since there is no previous work proposing alignment constraint for NAR, we design a new alignment constraint method and explore its effectiveness. We first calculate the expectation position of teacher's attention distributions for $i$-th target token: $E_{i} = \sum_{j=1}^{M}j*A'_{i,j}$ and cast it to the nearest integer. Then we constrain the attention weights of $i$-th target token for NAR model so that it can only attend to the source position between $E_{i-1}$ and $E_{i+1}$. Specially, the first target token can only attend to the source position between 1 and $E_{2}$ while the last target token can only attend to the position between $E_{N-1}$ and $M$. We apply this alignment constraint for ASR only in the training stage.

% \begin{table}[!h]
% \small
% 	\centering
% 	\begin{tabular}{ l | c | c | c  }
% 		\toprule
% 		Task & NAR & NAR + IAC & NAR + EAC \\
% 		\midrule
% 		NMT (BLEU) & 19.24 & 21.69 & / \\
% 		TTS (MOS)  & 1.06 \pm 0.23 & 1.97 \pm 0.16 & 3.61 \pm 0.13 \\
% 		ASR (WER)  & 36.2 & 33.1 & 21.5 \\
% 		\bottomrule
% 	\end{tabular}
% 	\caption{The results for attention constraint.}
% 	\label{tab_ac_results}
% \end{table}

\begin{table}[!h]
\small
	\centering
	\begin{tabular}{ l | c | c }
		\toprule
		Task & Model & Accuracy \\
		\midrule
		\multirow{3}{*}{NMT (BLEU) } & \textit{Transformer} & 33.90 \\
		\cmidrule{2-3}
		             & \textit{NAT} & 27.12   \\
		             & \textit{NAT w/o AC} & 25.03  \\
         \midrule 
		\multirow{3}{*}{ASR (WER)}  & \textit{Transformer ASR} & 20.1 \\
		\cmidrule{2-3}
		            & \textit{NAR-ASR} & 33.1 \\
		            & \textit{NAR-ASR w/o AC} & 39.23 \\
		\midrule 
		\multirow{3}{*}{TTS (MOS)}  & \textit{Transformer TTS} & 3.82 $\pm$ 0.08 \\
		\cmidrule{2-3}
		            & \textit{FastSpeech} & 3.79 $\pm$ 0.12 \\
		            & \textit{FastSpeech w/o AC} & 1.97 $\pm$ 0.16 \\
		\bottomrule
	\end{tabular}
	\caption{The comparison between NAR models with and without alignment constraint.}
	\label{tab_ac_results}
\end{table}

\paragraph{Experimental Results} We follow the model configuration and datasets as described in Section~\ref{gap_exp_settings}, and explore the accuracy improvements when adding attention constraint to NAR models. The results are shown in Table~\ref{tab_ac_results}. It can be seen that attention constraint can not only improve the performance of NMT and TTS as previous works~\citep{li2019hint,ren2019fastspeech} demonstrated, but also help the NAR-ASR model achieve better scores.

\begin{figure}[h]
	\centering
	\includegraphics[width=0.48\textwidth]{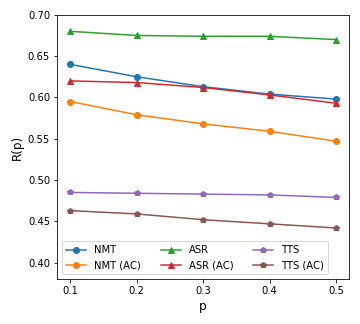}
	\caption{Attention density ratio $R(p)$ for NMT, ASR and TTS tasks under different $p$ with and without alignment constraint (AC).}
	\label{fig_adp_iac_res}
\end{figure}

\subsection{Why Alignment Constraint Works}
We further analyze how alignment constraint could help on NAR models by measuring the changes of token dependency when adding alignment constraint on CoMMA. 

For simplicity, we use the method described in Equation~\ref{eqn_iac} to help the training of CoMMA, where the teacher model is the AR model and student model is CoMMA. We minimize KL-divergence between the per-head encoder-to-decoder attention distributions of the AR model and CoMMA. First, we normalize the encoder-to-decoder attention weights in each head of mix-attention to convert each row of the attention weights to a distribution: 
\begin{equation}
\begin{aligned}
    \hat{A}_{i,j} = \frac{A_{i,N+j}}{\sum_{k=1}^{M}{A_{i,N+k}}} \\
    \textit{for each } i \in [1,N], j \in [1,M], 
\end{aligned} 
\end{equation}
where $A \subset \mathbb{R}^{N \times (N+M)} $ is the weights of mix-attention described in Section \ref{sec_ratio}, $\hat{A} \subset \mathbb{R}^{N \times M} $ is the normalized encoder-to-decoder attention weights, M and N is the length of source and target sequence. Then, we compute the KL-divergence loss for each head as follows:
\begin{equation} \mathcal{L}_{ac} = \frac{1}{N}\sum_{i=1}^{N}{D_{KL}(A'_{i}||\hat{A}_{i})},\end{equation}
where $A' \subset \mathbb{R}^{N \times M} $ is the encoder-to-decoder attention of AR teacher model. We average $L_{ac}$ over all heads and layers and get the final attention constraint loss for CoMMA.

We measure the token dependency by calculating the attention density ratio $R(p)$ based on CoMMA, and show the results in Figure~\ref{fig_adp_iac_res}. It can be seen that alignment constraint can help reduce ratio $R(p)$ on each task and thus reduce the dependency on target context when predicting target tokens. In the meanwhile, alignment constraint can help the model extract more information from the source context, which can help the learning of NAR models.

Another interesting finding is that NAR model in TTS benefits from attention constraint most as shown in Table~\ref{tab_ac_results}, and in the meanwhile, TTS has the least attention density ratio as shown in Figure~\ref{fig_adp_iac_res}. These observations suggest that NAR models with small target token dependency could benefit largely from alignment constraint.

% \begin{table}[h]
% \small
% 	\centering
% 	\begin{tabular}{ l | c | c | c  }
% 		\toprule
% 		Tasks          &  p=0.1  & p=0.3  & p=0.5 \\
% 		\midrule
% 		\textit{NMT}           & 0.640 & 0.613 & 0.598 \\
% 		\textit{NMT + IAC}      & 0.595 & 0.568 & 0.547 \\
% 		\midrule    
% 		\textit{ASR}           & 0.680  & 0.674 & 0.670 \\
% 		\textit{ASR + IAC}      & 0.540 & 0.518 & 0.502 \\
% 		\midrule      
% 		\textit{TTS}          & 0.485 & 0.483 & 0.479 \\
% 		\textit{TTS + IAC}     & 0.463 & 0.452 & 0.442 \\
% 		\bottomrule
% 	\end{tabular}
% 	\caption{Attention density ratio $R(p)$ for different tasks on different $p$}
% 	\label{tab_adp_kd_results}
% \end{table}

%We also need to modify CoMMA in order to study attention constraint. 

% For hard attention constraint, we take $A'$ as mask and apply it to the decoder-to-encoder attention submatrix $A_{1:M,1:N}$, which 

% For some low-$R$ tasks such as TTS, source information is extremely important for NAT model. A elaborately designed alignment constraint mechanism can be useful in these tasks, which is consistent with the previous work~\citep{ren2019fastspeech}.

\section{Related Works}
Several works try to analyze and understand NAR models on different tasks. We discuss these analyses from the two aspects: knowledge distillation and source-target alignment constraint.

\paragraph{Knowledge Distillation}
Knowledge distillation has long been used to compress the model size~\citep{hinton2015distilling,furlanello2018born,yang2018knowledge,anil2018large,li2017learning} or transfer the knowledge of teacher model to student model~\citep{tan2019multilingual,liu2019improving,liu2019end}, and soon been applied to NAR models~\citep{gu2017non,oord2017parallel,guo2019non,wang2019non,li2019hint,guo2019fine,ren2019fastspeech} to boost the accuracy. Some works focus on studying why knowledge distillation works: ~\citet{phuong2019towards} provide some insights into the mechanisms of knowledge distillation by studying the special case of linear and deep linear classifiers and find that data geometry, optimization bias and strong monotonicity determine the success of distillation; ~\citet{yuan2019revisit} argue that the success of KD is also due to the regularization of soft targets, which might be as important as the similarity information between categories.

However, few works have studied the cause of why knowledge distillation benefits NAR training. Recently, ~\citet{zhou2019understanding} investigate why knowledge distillation is important for the training of NAR model in NMT task and find that knowledge distillation can reduce the complexity of data sets and help NAR model to learn the variations in the output data. 
% However, these explanations are reasonable on its own, but mainly from the perspective of data level and is not easy to understand. Instead, we analyze knowledge distillation from a more understandable and intuitive perspective, by observing the change of the token dependency based on our proposed CoMMA.

~\citet{li2019hint} explore the causes of the poor performance of the NAR model by observing the attention distributions and hidden states of NAR model. \citet{lee2018deterministic} presents some experiments and analysis to prove the necessity for multiple iterations generation for NAT. They also investigate the effectiveness of knowledge distillation in different task and make the assumption that teacher model can essentially clean the training data so that the distilled NAR model substantially outperforms NAR model trained with raw data.

%Knowledge distillation was originally proposed for improving the performance of a student model with small model capacity by training it on the targets predicted from a strong teacher model. Knowledge distillation has soon been applied to a variety of tasks, including image classification~\citep{hinton2015distilling,furlanello2018born,yang2018knowledge,anil2018large,li2017learning}, speech recognition~\citep{hinton2015distilling} and natural language processing~\citep{DBLP:conf/emnlp/KimR16,freitag2017ensemble}.

%Recent works~\citep{furlanello2018born,yang2018knowledge} even demonstrate that student model can surpass the accuracy of the teacher model, even if the teacher model is of the same capacity as the student model. \citet{anil2018large} propose online distillation to improve the scalability of distributed model training and the training accuracy. Some researchers focus on transfering the knowledge from the teacher to student model in multi-task situations~\citep{liu2019improving,tan2019multilingual} and achieve great performance. 

%Some works focus on transferring knowledge from teacher to student of different tasks but with same model inputs and outputs. For example, \citet{liu2019end} transfer the knowledge from text to text neural machine translation (NMT) and automatic speech recognition (ASR) teacher to end-to-end speech to text translation student. 

\paragraph{Attention Alignment Constraint}
Previous work pointed out that adding additional alignment knowledge can improve the estimation of the soft alignment in attention mechanism model. For example, \citet{DBLP:journals/corr/ChenMKP16} uses the Viterbi alignments of the IBM model 4 as an additional knowledge during NMT training by calculating the divergence between the attention weights and the statistical alignment information.

Compared with AR model, the attention distributions of NAR model are more ambiguous, which leads to the poor performance of the NAR model. Recent works employ attention alignment constraint between the well-trained AR and NAR model to train a better NAR model. \citet{li2019hint} leverages intermediate hidden information from a well-trained AR-NMT teacher model to improve the NAR-NMT model by minimizing KL-divergence between the per-head encoder-decoder attention of the teacher and the student. \citet{ren2019fastspeech} choose the encoder-decoder attention head from the AR-TTS teacher as the attention alignments to improve the performance of the NAR model in TTS.

\section{Conclusion}
In this paper, we conducted a comprehensive study on NAR models in NMT, ASR and TTS tasks to analyze several research questions, including the difficulty of NAR generation and why knowledge distillation and alignment constraint can help NAR models. We design a novel CoMMA and a metric called attention density ratio to measure the dependency on target context when predicting a target token, which can analyze these questions in a unified method. Through a series of empirical studies, we demonstrate that the difficulty of NAR generation correlates on the target token dependency, and knowledge distillation as well as alignment constraint reduces the dependency of target tokens and encourages the model to rely more on source context for target token prediction, which improves the accuracy of NAR models. We believe our analyses can shed light on the understandings and further improvements on NAR models.

\section*{Acknowledgments}

This work was supported in part by the National Key R\&D Program of China (Grant No.2018AAA0100603), Zhejiang Natural Science Foundation (LR19F020006), National Natural Science Foundation of China (Grant No.61836002), National Natural Science Foundation of China (Grant No.U1611461), and National Natural Science Foundation of China (Grant No.61751209). This work was also partially funded by Microsoft Research Asia. Thanks Tao Qin for the valuable suggestions, comments and guidance on this paper.

\bibliography{acl2020}
\bibliographystyle{acl_natbib}

\clearpage

\onecolumn

\section*{Supplementary Material}
\setcounter{section}{0}
\section{Model Settings of NAR and AR}
We show the model settings of NAR and AR in Table \ref{tab:ar_nar_hyperparameters}. The hyperpameters in pre-net follow the methods in each task listed in Table \ref{ar_nar_model} in the main part of the paper.

\begin{table}[!h]
\centering
\begin{tabular}{l|l|l|l}
\hline
\textbf{Transformer Hyperparameter} & \textbf{NMT / NAT} & \textbf{ASR / NAR-ASR} & \textbf{TTS / FastSpeech}  \\           \hline\hline
Embedding Dimension     & 512  & 512   & 512        \\ \hline
Encoder Layers          & 6    & 6     & 6           \\ \hline
Encoder Hidden          & 512  & 512   & 512         \\ \hline
Encoder Filter Size     & 1024 & 1024  & 1024             \\ \hline
Encoder Heads           & 4    & 4     & 4            \\ \hline
Decoder Layers          & 6    & 6     & 6          \\\hline
Decoder Hidden Size     & 512  & 512   & 512            \\ \hline
Decoder Filter Size     & 1024 & 1024  & 1024           \\ \hline
Decoder Heads           & 4    & 4     & 4                   \\ \hline
Dropout                 & 0.2  & 0.1   & 0.2             \\   \hline
Batch Size              & 64   & 32    & 32        \\ \hline 
Base Learning Rate      & 1e-3 & 1e-3 & 1e-3  \\ \hline 
\end{tabular}
\caption{Hyperparameters of transformer-based AR and NAR models.}
\label{tab:ar_nar_hyperparameters}
\end{table}

\section{Model Settings of CoMMA}
We show the model settings of CoMMA in Table \ref{tab:comma_hyperparameters}.

\begin{table}[!h]
\centering
\begin{tabular}{l|l}
\hline
\textbf{Name} & \textbf{Hyperparameter}     \\ \hline\hline
Embedding Dimension     & 512      \\ \hline
Encoder Layers          & 6         \\ \hline
Encoder Hidden          & 512       \\ \hline
Encoder Filter Size     & 1024           \\ \hline
Encoder Heads           & 4          \\ \hline
Decoder Layers          & 6                    \\ \hline
Decoder Hidden Size     & 512          \\ \hline
Decoder Filter Size     & 1024         \\ \hline
Decoder Heads           & 4                 \\ \hline
Dropout                 & 0.1           \\   \hline
Batch Size              & 64      \\ \hline 
Base Learning Rate      & 1e-3  \\ \hline 

\end{tabular}
\caption{Hyperparameters of CoMMA.}
\label{tab:comma_hyperparameters}
\end{table}

\end{document}